\documentclass[]{bytedance_seed}
\usepackage{helvet}
\usepackage{amsmath} 
\usepackage{natbib}
\usepackage{graphicx}
\usepackage{subcaption}

\usepackage[toc,page,header]{appendix}
\usepackage[utf8]{inputenc} % allow utf-8 input
\usepackage[T1]{fontenc}    % use 8-bit T1 fonts
\usepackage{hyperref}       % hyperlinks
\usepackage{url}            % simple URL typesetting
\usepackage{booktabs}       % professional-quality tables
\usepackage{lmodern}        % scalable Latin Modern fonts
\usepackage{amsfonts}       % blackboard math symbols
\usepackage{nicefrac}       % compact symbols for 1/2, etc.
\usepackage{microtype}      % microtypography
\usepackage{wrapfig} 

\usepackage{amssymb}  % 用于特殊符号如♠♣等
\usepackage{fontawesome}  % 用于邮箱图标 \faEnvelope
\usepackage{url}  % 用于 \url 命令
\usepackage{minitoc}

\usepackage{enumitem}
\usepackage{tablefootnote}
\usepackage{dsfont}
\usepackage[flushleft]{threeparttable}
\usepackage{multirow}
\usepackage[svgnames]{xcolor}
\usepackage[most]{tcolorbox}
\usepackage{caption}
\usepackage{subcaption}

\usepackage{amssymb}
\usepackage{bbding}
\usepackage{float}

\usepackage{pgfplots}
\pgfplotsset{compat=1.18}

\usepackage[table]{xcolor} % enables color support for tables
\usepackage{colortbl}      % provides \rowcolor and \columncolor

\usepackage{booktabs}
\usepackage{array}
\usepackage{etoolbox}

\definecolor{lightblue}{RGB}{200, 230, 255}  
\definecolor{headerblue}{RGB}{150, 200, 255} 

\definecolor{Gray}{gray}{0.8}

\title{Mini-o3: Scaling Up Reasoning Patterns and Interaction Turns for Visual Search}
\author[1*\dagger]{Xin Lai}
\author[1,2*]{Junyi Li}
\author[1]{Wei Li}
\author[1]{Tao Liu}
\author[1]{Tianjian Li}
\author[2\dagger]{Hengshuang Zhao}

\affiliation[1]{ByteDance}
\affiliation[2]{The University of Hong Kong}

\contribution[*]{Equal contribution}
\contribution[\dagger]{Corresponding authors}

\abstract{
\begin{abstract}

Recent advances in large multimodal models have leveraged image-based tools with reinforcement learning to tackle visual problems. However, existing open-source approaches often exhibit monotonous reasoning patterns and allow only a limited number of interaction turns, making them inadequate for difficult tasks that require trial-and-error exploration. In this work, we address this limitation by scaling up tool-based interactions and introduce Mini-o3, a system that executes deep, multi-turn reasoning—spanning tens of steps—and achieves state-of-the-art performance on challenging visual search tasks. Our recipe for reproducing OpenAI o3–style behaviors comprises three key components. First, we construct the Visual Probe Dataset, a collection of thousands of challenging visual search problems designed for exploratory reasoning. Second, we develop an iterative data collection pipeline to obtain cold-start trajectories that exhibit diverse reasoning patterns, including depth-first search, trial-and-error, and goal maintenance. Third, we propose an over-turn masking strategy that prevents penalization of over-turn responses (those that hit the maximum number of turns) during reinforcement learning, thereby balancing training-time efficiency with test-time scalability. Despite training with an upper bound of only six interaction turns, our model generates trajectories that naturally scale to tens of turns at inference time, with accuracy improving as the number of turns increases. Extensive experiments demonstrate that Mini-o3 produces rich reasoning patterns and deep thinking paths, effectively solving challenging visual search problems.
\end{abstract}
}

\date{\today}
\correspondence{Xin Lai at \email{laixin.1@bytedance.com}, Hengshuang Zhao at \email{hszhao@cs.hku.hk}}

\checkdata[Project Page]{\url{https://github.com/Mini-o3/Mini-o3}}

\begin{document}
\maketitle

%不需要目录就注释掉 注意目录不要和第一页放在一块 要有\newpage
%\newpage
%\tableofcontents
%\newpage

% \input{sections/introduction}
% \input{sections/relatedwork}
% \input{sections/preliminary}
% \input{sections/method}
% \input{sections/experiments}
% \input{sections/conclusion}

\begin{figure}[h]
  \centering
  % 每个子图占栏宽的0.48，保证并排，两者之间留少量空隙
  \begin{subfigure}{0.47\linewidth}
    \centering
\begin{tikzpicture}
      \begin{axis}[
        title={Accuracy on VisualProbe-Hard},
        width=\linewidth,
        height=1.0\linewidth,
        symbolic x coords={4,8,16,24,32},
        xtick=data,
        ymin=24, ymax=59,
        ytick={24,26,28,30,32,34,36,38,40,42,44,46,48,50,52,54,56,58,60},
        xlabel={Upper Limit of Turns During Testing},
        % 关闭默认ylabel
        ylabel={},
        % 留出上边距，避免顶到边框
        clip=false,
        grid=both,
        grid style={dashed, gray!30},
        major grid style={dashed, gray!45},
        minor y tick num=1,
        legend style={at={(0.02,0.98)}, anchor=north west, draw=none, fill=white, fill opacity=0.85},
        legend cell align=left,
        tick label style={/pgf/number format/fixed},
        line width=1.1pt,
        mark size=2.3pt,
      ]
    % 数据
    \addplot+[color=green!60!black, mark=*]
      coordinates {(4,25.3) (8,40.6) (16,47.0) (24,47.8) (32,48.0)};
    \addlegendentry{Mini-o3 (Ours)}
    \addplot+[color=red!70!black, mark=square*]
      coordinates {(4,38.1) (8,38.3) (16,38.3) (24,38.3) (32,38.3)};
    \addlegendentry{Mini-o3 w/o over-turn mask}
    \addplot+[color=blue!70!black, mark=triangle*]
      coordinates {(4,35.0) (8,35.1) (16,35.1) (24,35.1) (32,35.1)};
    \addlegendentry{DeepEyes}
    % 在纵轴刻度数字的正上方添加“avg@32”
    % axis description cs: (0,1) 是左上角；向下微调一点以紧贴刻度上方
    \node[anchor=south west, font=\normalsize]
      at (axis description cs:-0.18,1.01) {Avg@32};
      \end{axis}
\end{tikzpicture}
  \end{subfigure}\hfill
  \begin{subfigure}{0.5\linewidth}
    \centering
\includegraphics[width=0.99\linewidth]{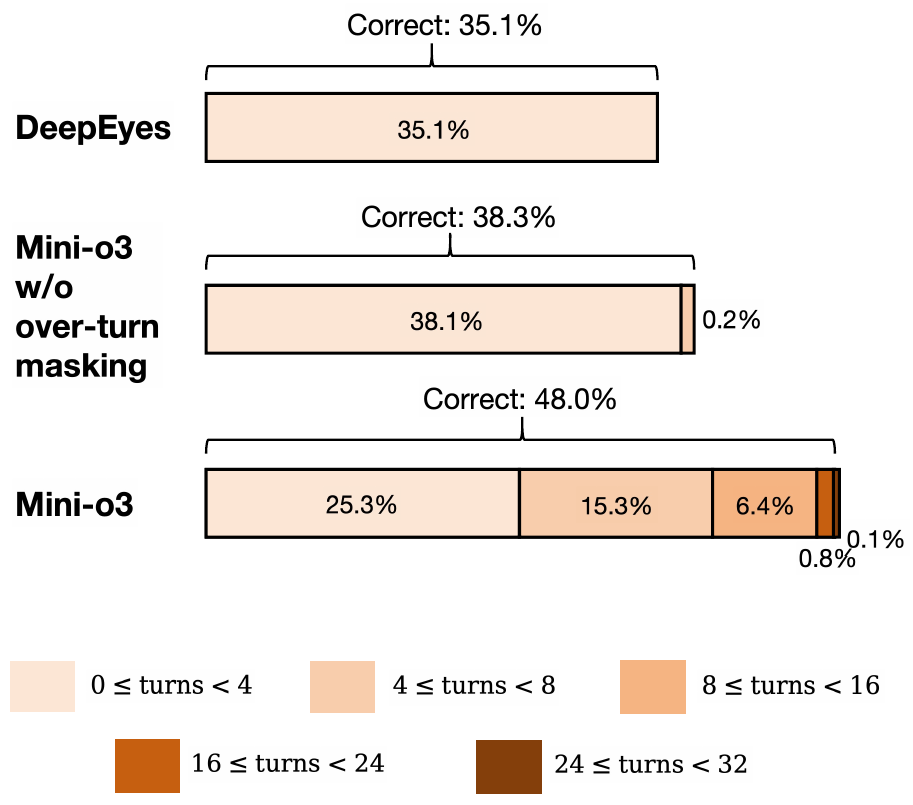}
\vspace{0.3cm}
  \end{subfigure}
  % \vspace{-0.5cm}
  \caption{\textbf{Left}: Visual search accuracy continues to grow as the upper limit on the number of turns increases for Mini-o3. \textbf{Right}: Distribution of the correct trajectories under different numbers of interaction turns during testing. Mini-o3 demonstrates deeper thinking paths and stronger performance. Despite a small upper limit (i.e., $6$ turns) during training, it shows the test-time turns scaling property: accuracy continues to grow as the maximum number of turns increases from $4$ to $32$.}
  \label{fig:intro}
\end{figure}
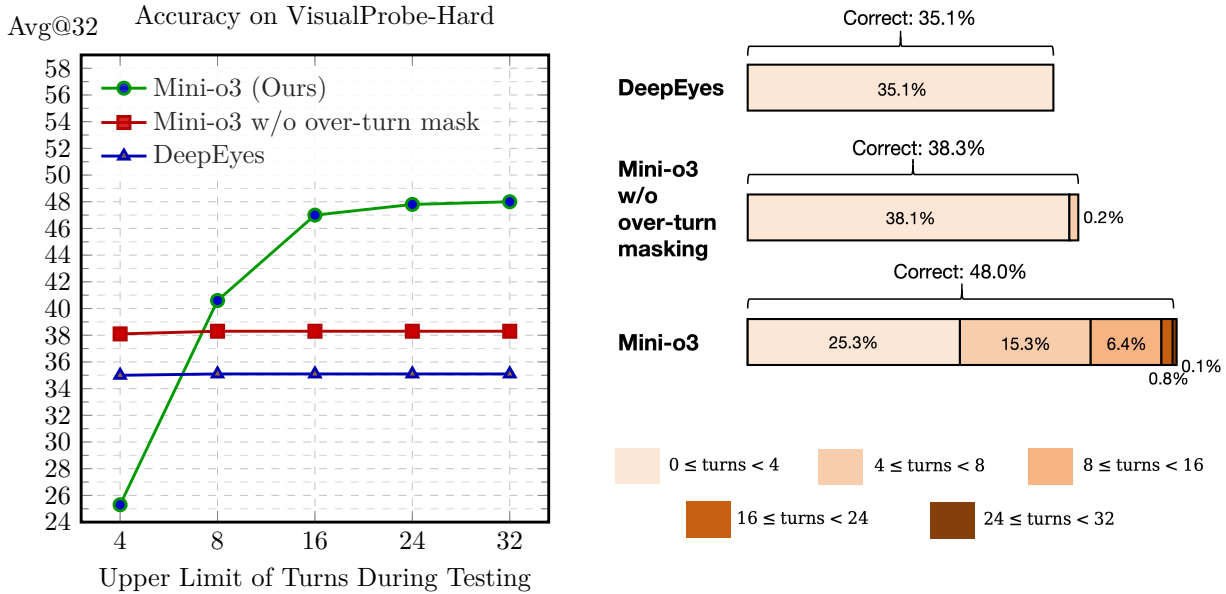

\section{Introduction}

\begin{figure}[p]
\centering
\includegraphics[width=0.99\linewidth]{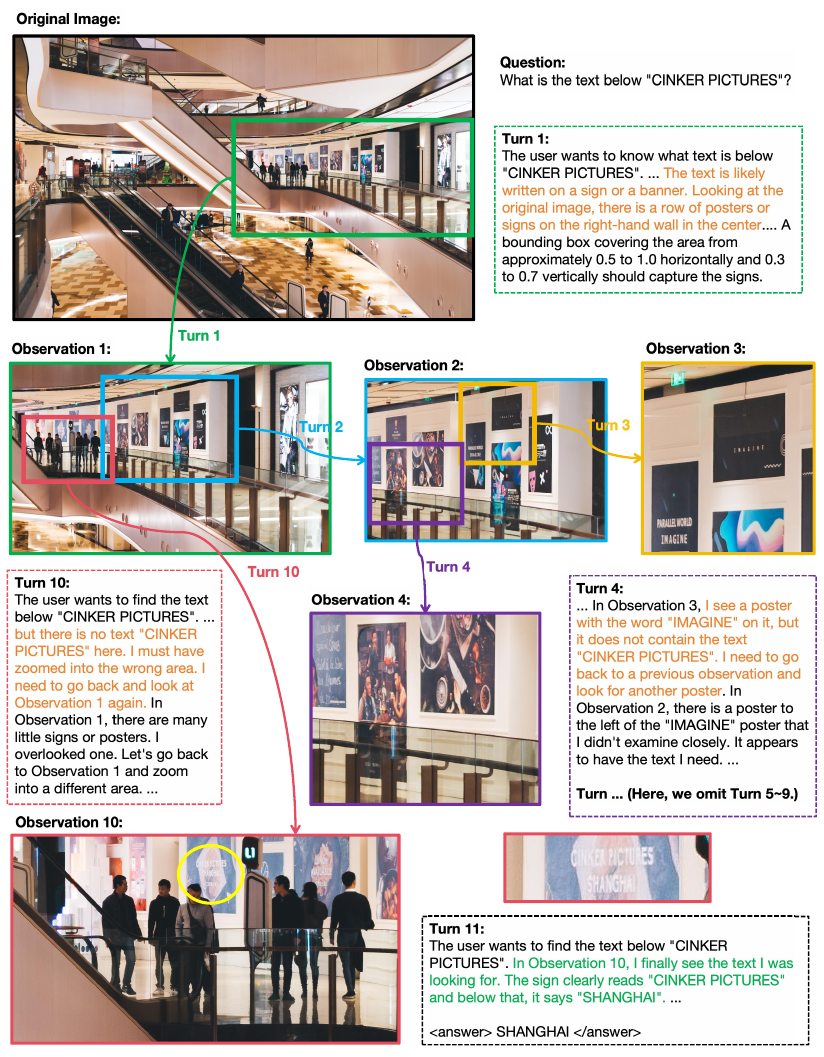}
\vspace{-0.3cm}
\caption{A multi-turn trajectory generated by Mini-o3. It shows complicated reasoning patterns (e.g., trial-and-error exploration) and deep thinking paths (i.e., 11 turns) in visual search tasks. More illustrations are given in Appendix.}
\label{fig:demo}
\end{figure}

Recently, the capability to invoke image-centric tools has been incorporated into a wide range of Vision–Language Models (VLMs). This thinking-with-image capability enables flexible visual operations and fine-grained reasoning, substantially advancing visual understanding.

However, while existing open-source VLMs exhibit solid performance on relatively simple visual search benchmarks (e.g., V\textsuperscript{*} Bench~\citep{wu2024v}, HR-Bench~\citep{wang2025divide}), they remain weak on challenging tasks that require trial-and-error exploration. As shown in Fig.\ref{fig:intro}, DeepEyes\citep{zheng2025deepeyes} achieves only $35.1\%$ accuracy on VisualProbe-Hard. We further observe that this underperformance on difficult problems stems from monotonous reasoning patterns and limited interaction turns. For instance, in HR-Bench-4K, DeepEyes uses image tools for an average of merely one turn per example. Unlike OpenAI o3~\citep{o3}, these models fail to produce diverse reasoning strategies (e.g., depth-first search, trial-and-error exploration, self-reflection) and deep thinking trajectories spanning tens of tool-interaction rounds.

Motivated by these observations, we present Mini-o3 and provide a complete recipe to reproduce the thinking-with-image capability with behaviors similar to OpenAI o3. As illustrated in Fig.\ref{fig:demo}, Mini-o3 generates complex reasoning patterns and deep interaction trajectories, delivering unprecedented performance on challenging visual search tasks. Moreover, Fig.\ref{fig:intro} (left) demonstrates Mini-o3’s ability to scale the number of interaction turns at test time: accuracy consistently improves as the upper bound on interaction turns increases from 4 to 32 during inference, despite training with a budget of only 6 turns. By scaling both the depth of interaction and the diversity of reasoning patterns, Mini-o3 expands the solvable frontier of difficult problems, as shown in Fig.~\ref{fig:intro} (right).

Our training recipe comprises three components. First, we construct the Visual Probe Dataset, which contains thousands of high-resolution images paired with challenging visual search questions and answers. In contrast to prior benchmarks (e.g., V\textsuperscript{*} Bench, HR-Bench), where targets are often easy to localize, our problems are explicitly designed to require trial-and-error exploration. Notably, the inclusion of such challenging training samples is essential to elicit diverse reasoning patterns and deep interaction trajectories under reinforcement learning.

Second, we develop an effective pipeline to iteratively synthesize diverse multi-turn trajectories for cold-start supervised finetuning. Concretely, we begin by crafting a small set of representative demonstrations, each comprising the input image and question, along with per-turn observations, thoughts, and actions. These demonstrations cover varied reasoning strategies, including depth-first search, self-reflection, and goal maintenance. We then prompt an existing VLM to mimic these behaviors in a few-shot manner and to produce the thought and action for each turn on new queries, iterating until the model completes the task or reaches the interaction budget. Only trajectories that culminate in a correct answer are retained. Importantly, the base VLM used for data synthesis need not possess native thinking-with-image ability; in-context mimicking suffices.

Third, to enable scaling the number of interaction turns at inference time for harder problems, we avoid penalizing the over-turn trajectories (those that exceed the upper limit of interaction turns) and introduce an over-turn masking technique in reinforcement learning. Specifically, we mask advantages for trajectories that hit the upper limit of interaction turns or the context length. Consequently, over-turn trajectories are ignored during policy updates, and their losses do not contribute gradients. This simple yet effective strategy encourages the emergence of more complex reasoning patterns without overfitting to short trajectories, thereby supporting test-time scaling of interaction depth. It also alleviates the need for a large training-time turn budget: in our experiments, we cap training at only 6 turns, significantly improving efficiency. For example, reducing the training budget from 16 to 6 turns shortens a 10-day training run to about 3 days, with negligible impact on test accuracy.

\begin{figure}[t]
\centering
\includegraphics[width=1.0\linewidth]{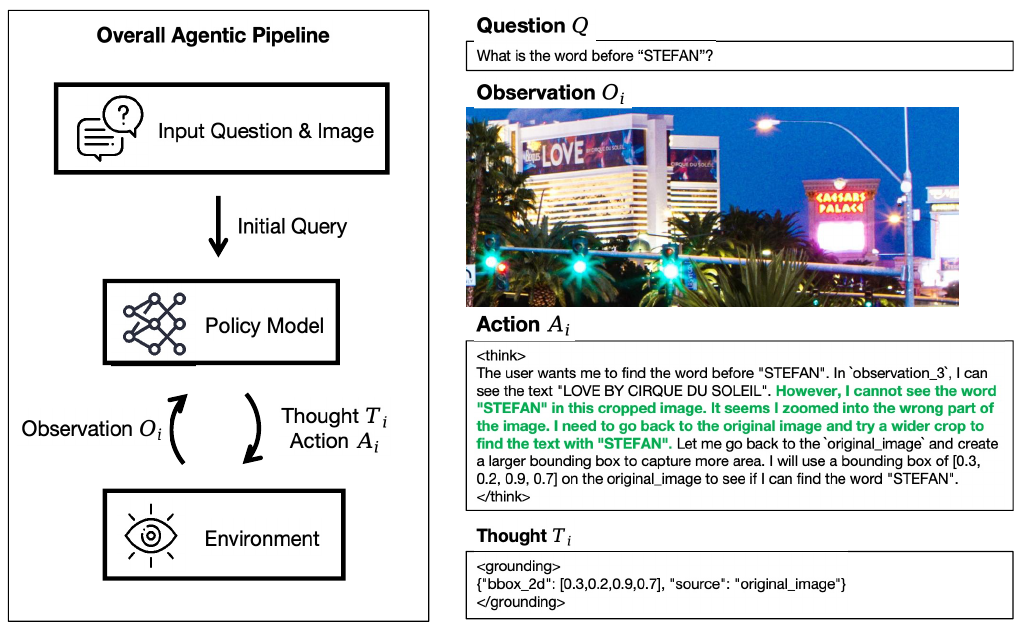}
\caption{The overview of our framework for multi-turn agentic image tool use. During each turn, the model generates the thought and action iteratively based on the previous observation (or the input question and image). The observation at each turn is obtained based on the parameters indicated by the corresponding action.}
\label{fig:overview}
\end{figure}

\section{Related Work}
\subsection{Vision-Language Models}
The emergence of Vision-Language Models (VLMs) has marked a major milestone in artificial intelligence by enabling the joint understanding of visual and textual modalities. Early seminal works, including BLIP-2~\citep{li2023blip}, Flamingo~\citep{alayrac2022flamingo}, and the LLaVA series~\citep{liu2024visual,li2024llava,guo2024llava}, established a foundational paradigm that couples strong pre-trained vision encoders (e.g., ViT~\citep{dosovitskiy2020image}) with large language models (LLMs). These systems typically introduce a projector to align visual features with the linguistic embedding space, thereby endowing LLMs with visual grounding.
Building on this paradigm, more recent multimodal models—such as Gemini~\citep{gemini}, GPT-4o~\citep{gpt4o}, and Qwen2.5-VL~\citep{bai2025qwen2.5}, among others~\citep{claude-3.5,llama-3.2,aira,internvl,vila}—have achieved state-of-the-art performance on a wide range of visual understanding tasks, notably visual question answering. Their gains are largely driven by scaling model capacity and training on diverse, high-quality image–text corpora.
In parallel, advances in reinforcement learning have enhanced the reasoning capabilities of VLMs by encouraging structured, step-by-step problem solving via Chain-of-Thought prompting~\citep{wei2022chain}. Recent approaches~\citep{meng2025mmeureka,huang2025vision,shao2024visual,liu2025visual,zhang2025r1,zhou2025r1} primarily target improved textual reasoning for challenging tasks, including counting, logical inference, and mathematical problem solving.

\subsection{Tool-Integrated Agents with Reinforcement Learning}
Progress in reinforcement learning (RL) including algorithms such as REINFORCE~\citep{reinforce}, PPO~\citep{ppo}, RLOO~\citep{rloo}, ReMax~\citep{remax}, GRPO~\cite{deepseekmath}, REINFORCE++\citep{reinforce++}, Dr.GRPO\citep{drgrpo}, and GSPO~\citep{GSPO} has substantially reshaped training paradigms for both LLMs and VLMs. Systems like DeepSeek-R1~\citep{guo2025deepseek} and Kimi-K1.5~\citep{kimi-k1.5} further demonstrated the efficacy of simple, verifiable reward signals in RL for improving reasoning quality.
More recently, tool-augmented agents—such as OpenAI’s o3 and o4~\citep{o3}, Kimi-Researcher~\citep{kimi-researcher}, Kimi-K2~\citep{kimi-k2}, and others~\citep{tao2025webshaper,geng2025webwatcher,li2025websailor,mai2025agent,simpletir}—have shown strong agentic abilities in long-horizon, multi-turn tasks by leveraging a broad toolkit (e.g., web browsing, code execution, retrieval). Complementary lines of work, including DeepEyes~\citep{zheng2025deepeyes}, Chain-of-Focus~\citep{zhang2025chain}, and Pixel Reasoner~\citep{su2025pixel}, as well as related methods~\citep{zhu2025active,yang2025visionthink,wu2025mmsearch,huang2025high}, aim to equip VLMs with iterative zoom-in and region-of-interest selection, enabling active perception over images.
While these directions collectively point to a promising path for next-generation visual understanding --- particularly on challenging, compositional problems --- current models often exhibit limited interaction depth and overly rigid reasoning patterns, constraining their effectiveness in complex settings. Our work advances this line by presenting an effective training recipe for a multimodal agent that supports multi-turn image tool use, thereby improving adaptability and reasoning diversity in visually grounded tasks.

% \section{Preliminary}

\section{Our Approach}

\subsection{Overview}

\paragraph{Overall Agentic Pipeline} We illustrate the overall agentic pipeline in Fig.~\ref{fig:overview}. Given a user query and an input image, the policy model iteratively produces a thought $T_i$ and an action $A_i$. The action interacts with the environment by invoking image tools, which yields a new observation $O_i$. This observation is appended to the interaction history and fed back to the policy model. The thought–action–observation loop terminates when the model returns a final answer or when predefined limits on context length or interaction turns are reached. The components are detailed below.
% \vspace{-0.5cm}
\begin{itemize}
    \item Thought $T_i$: The internal reasoning process used by the policy model to select the next action, conditioned on the interaction history and the current observation. We encourage diverse reasoning patterns within thoughts to facilitate trial-and-error exploration for challenging problems.
    \item Action $A_i$: The action space comprises two options: (1) grounding and (2) emitting a final answer. For grounding, we parameterize the action with: \texttt{bbox\_2d}: The normalized bounding box in $[0,1]^2$ specifying the zoom-in region. \texttt{source}: The image on which the grounding operates, chosen from `original\_image'' or `observation\_i''. This design allows the model to act on any prior observation in the trajectory.
    \item Observation $O_i$: The observation produced by executing $A_i$ in the environment. Concretely, it is the image patch cropped either from the original image or from a historical observation.
\end{itemize}

\paragraph{Two-phase Training} Our training procedure consists of two phases.
\begin{itemize}
    \item Supervised Fine-Tuning (SFT): We first fine-tune the model on thousands of multi-turn trajectories involving image tool use (i.e., cold-start data). The objective is to teach the model to generate valid trajectories with diverse and robust reasoning patterns.
    \item Reinforcement Learning with Verifiable Rewards (RLVR): We then apply GRPO~\citep{deepseekmath} to optimize the policy with verifiable, semantics-aware rewards. Because many ground-truth answers in our RL data require semantic rather than exact string matching, we employ an external LLM as a judge to compute reward signals. To maintain training efficiency and stability, we impose upper bounds of 6 interaction turns and a 32K context length.
\end{itemize}

\subsection{Training Data Collection}

\begin{figure}[t]
\centering
\includegraphics[width=1.0\linewidth]{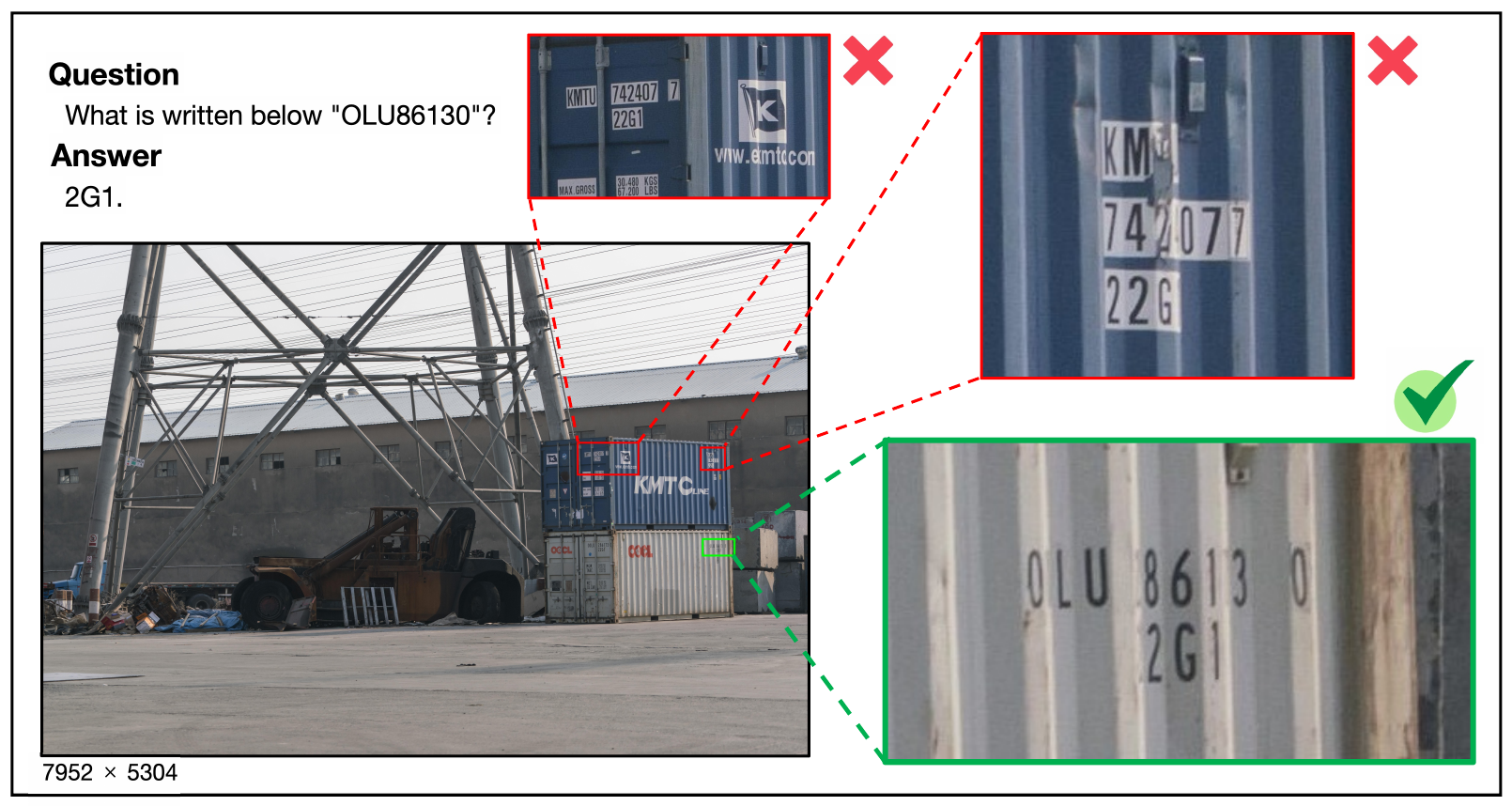}
\caption{Illustration of the VisualProbe dataset. The VisualProbe dataset features 1) small targets; 2) disturbance objects; 3) high-resolution images. As a result, it is super challenging and requires iterative exploration and trial-and-error.}
\label{fig:visual_probe_dataset}
\end{figure}

\paragraph{Visual Probe Dataset} Hard instances are essential for encouraging reflective, trial-and-error reasoning during reinforcement learning. To this end, we construct a challenging visual search dataset, the Visual Probe Dataset (VisualProbe). It comprises $4,000$ visual question–answer pairs for training and $500$ pairs for testing, spanning three difficulty levels: easy, medium, and hard. Compared with prior visual search benchmarks (e.g., V\textsuperscript{*} Bench), VisualProbe is characterized by: (1) small targets, (2) numerous distractor objects, and (3) high-resolution images, as illustrated in Fig.~\ref{fig:visual_probe_dataset}. These properties make the tasks substantially more demanding and naturally require iterative exploration and trial-and-error.

\begin{figure}[t]
\centering
\includegraphics[width=0.75\linewidth]{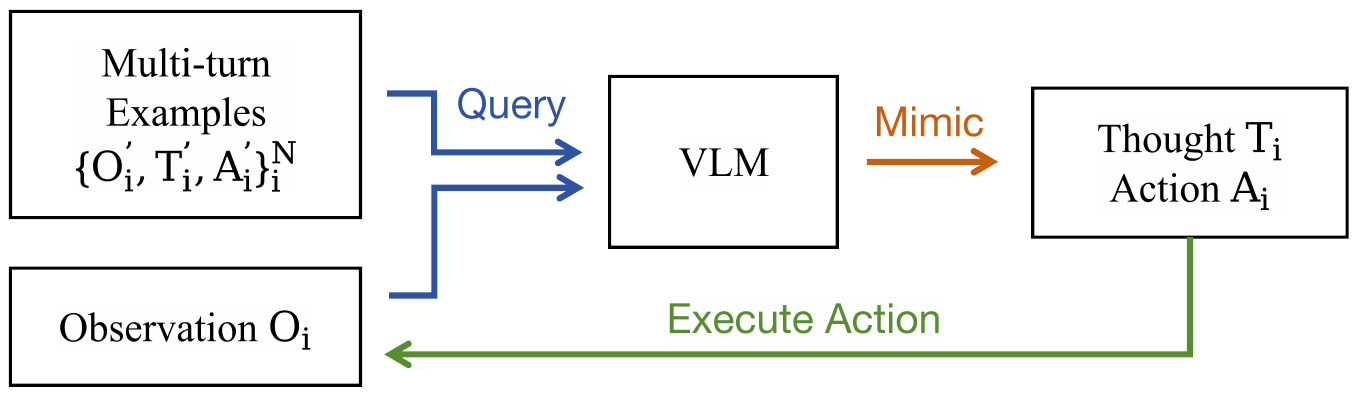}
\caption{The pipeline of cold-start data collection.}
\label{fig:cold_start}
\end{figure}

\paragraph{Diverse Cold-start Data} We initially attempted to train the model with reinforcement learning alone, without cold-start supervised fine-tuning (SFT). However, the model tended to produce concise responses and trajectories with few turns. We attribute this behavior to the base model’s lack of exposure to long-horizon agentic trajectories during pretraining and instruction tuning (here, \texttt{Qwen2.5-VL-7B-Instruct}). To handle complex exploratory tasks, we thus employ cold-start SFT to activate multi-turn tool-use capabilities.

The cold-start data collection pipeline is shown in Fig.~\ref{fig:cold_start}. To generate high-quality, diverse multi-turn trajectories, we prompt an existing VLM with in-context learning ability using a small set of manually crafted exemplars. The VLM is instructed to imitate the exemplars by iteratively producing a thought and an action at each turn. The loop terminates upon emitting a final answer or reaching a pre-defined turn limit. We retain only trajectories whose final answers are correct. Following this procedure, we collect approximately $6,000$ cold-start trajectories from $6$ exemplars.

\subsection{Reinforcement Learning}

\paragraph{Lower Down Max Pixels} The base model’s context length is constrained to 32K tokens. With the default image budget of roughly 12M pixels, the allowable number of interaction turns becomes severely limited by context, which hampers trial-and-error exploration on difficult tasks. To increase the feasible turn count per episode, we reduce the maximum pixels per image to 2M (or lower if necessary). This simple adjustment allows more turns to fit within the same context budget, improving solve rates on long-horizon problems. %\xinlai{add figure to compare the rate of solve\_acc\_none or pass@k}

\paragraph{Over-turn Masking} In the vanilla GRPO setting, each question $q$ is passed to the policy model to generate a group of outputs $\{o_i\}_{i=1}^{G}$. Rewards $r$ are then computed based on the correctness of the responses. Notably, when a response hits the maximum number of turns or exceeds the context length limit, the reward is set to $0$, as no valid answer can be produced in such cases. Subsequently, we compute advantages $A$ by normalizing the rewards and update the policy using the GRPO optimization objective over mini-batches. In our implementation, we do not include KL or entropy regularization. Formally, the optimization objective is given by:
\begin{equation}
\begin{split}
    \mathcal{J}_{GRPO}(\theta)  = \mathbb{E}_{[q \sim \mathcal{D}, \{o_i\}_{i=1}^G \sim \pi_{\theta_{old}}(\cdot|q)]} \frac{1}{G}\sum_{i=1}^G \left(\min\left(\frac{\pi_\theta(o_i |q)}{\pi_{\theta_{old}}(o_i |q)} A_i, \text{clip} \left( \frac{\pi_\theta(o_i |q)}{\pi_{\theta_{old}}(o_i |q)}, 1 - \epsilon, 1 + \epsilon \right)  A_i \right) \right)
\end{split}
\label{eq:GRPO}
\end{equation}
\begin{equation}
\begin{split}
    A_i = \frac{r_i - mean(\{r_1,r_2,...,r_G\})}{std(\{r_1, r_2, ...,r_G\})}.
\end{split}
\label{eq:advantage}
\end{equation}

\begin{figure}[t]
\centering
\begin{tikzpicture}
  \begin{axis}[
    width=14cm,
    height=6.4cm, % lowered overall height
    ybar,
    bar width=9pt,
    title style={yshift=4pt, font=\bfseries},
    ylabel={Percentage (\%)},
    ymin=0, ymax=75,
    enlarge x limits=0.15,
    symbolic x coords={
      {[1, 4)},
      {[4, 8)},
      {[8, 16)},
      {[16, 24)},
      {[24, 32)}
    },
    xtick=data,
    xticklabel style={font=\small},
    yticklabel style={font=\small},
    legend style={
      at={(0.98,0.98)}, anchor=north east,
      draw=none, fill=white, fill opacity=0.8, text opacity=1,
      font=\small
    },
    ymajorgrids=true,
    grid style={dashed,gray!40},
    nodes near coords,
    nodes near coords style={font=\scriptsize, /pgf/number format/.cd, fixed, precision=1},
    every axis plot/.append style={fill opacity=0.9},
  ]
    % Dataset 1: Mini-o3 w/o overlength masking
    \addplot+[draw=black, fill=blue!55, area legend, bar shift=-6pt] coordinates {
      ({[1, 4)}, 69.4)
      ({[4, 8)}, 29.5)
      ({[8, 16)}, 1.1)
      ({[16, 24)}, 0.0)
      ({[24, 32)}, 0.0)
    };
    % Dataset 2: Mini-o3
    \addplot+[draw=black, fill=orange!75, area legend, bar shift=6pt] coordinates {
      ({[1, 4)}, 53.3)
      ({[4, 8)}, 32.4)
      ({[8, 16)}, 12.2)
      ({[16, 24)}, 1.9)
      ({[24, 32)}, 0.1)
    };
    \legend{
      {Mini-o3 w/o over-turn masking},
      {Mini-o3}
    }
  \end{axis}
\end{tikzpicture}
\vspace{-0.3cm}
\caption{Distribution of interaction-turn percentages across five turn ranges during testing on VisualProbe-Hard. The percentages are calculated only on the \textit{correct} responses.}
\label{fig:turn_dist}
\end{figure}
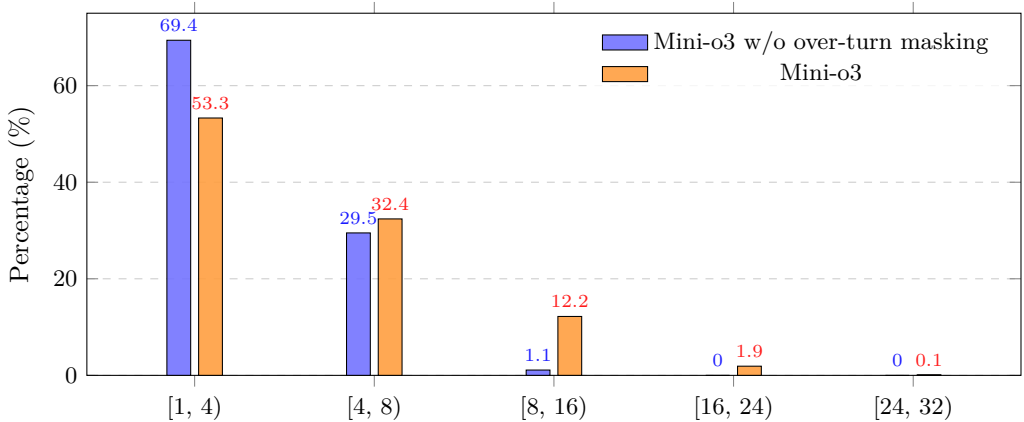

\begin{figure}[t]
\centering
\includegraphics[width=1.0\linewidth]{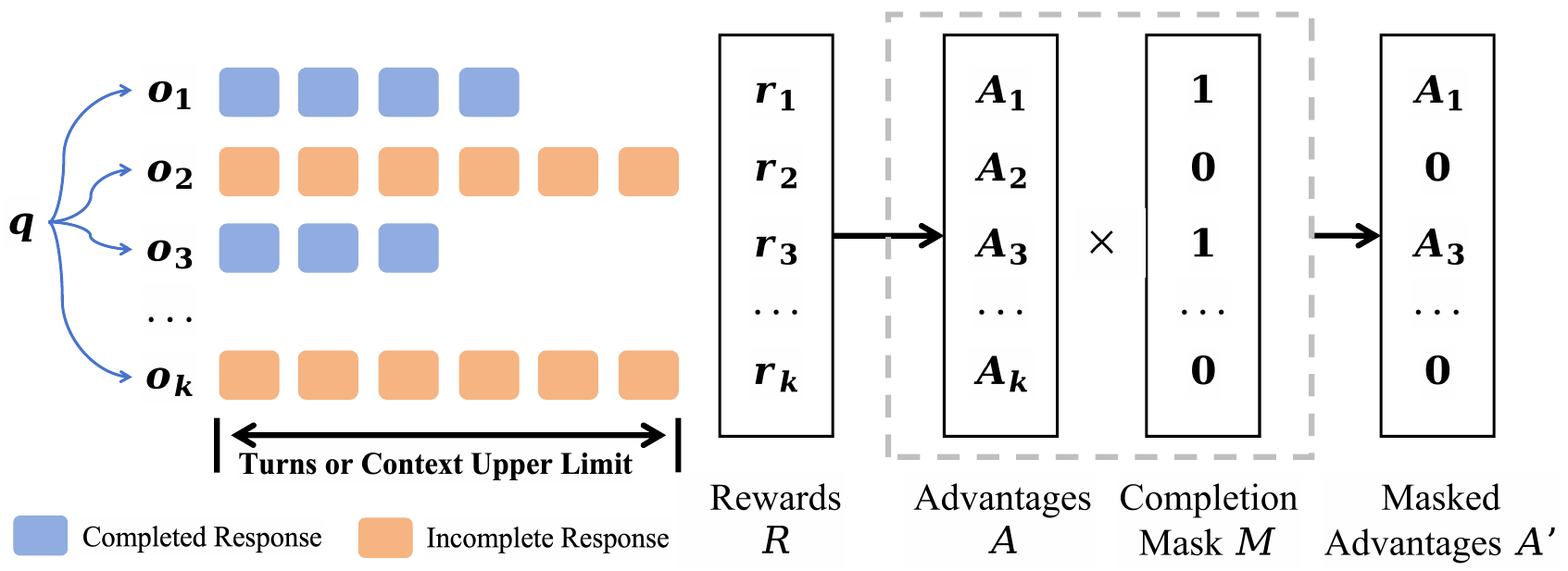}
\vspace{-0.3cm}
\caption{Illustration of the over-turn masking technique. The incomplete responses refer to those that exceed the maximum limit of interaction turns or context length.}
\label{fig:overlength_mask}
\end{figure}

However, we observe that over-turn responses --- those that hit the maximum number of turns or exceed the context length --- are assigned zero reward, which translates into negative advantages after normalization. In effect, such responses are penalized and discouraged throughout training.

This design has two drawbacks. First, the correctness of over-turn responses is inherently unknown; blunt penalization thus injects label noise into the return signal and can destabilize training. Second, for efficiency, the turn limit during training must remain modest (typically fewer than $10$ turns). As a consequence, over-turn responses occur frequently --- exceeding $20\%$ at the beginning of training. In this regime, naïve penalization biases the model to answer prematurely, substantially suppressing the number of interaction turns (see Fig.~\ref{fig:turn_dist}). This makes highly challenging tasks intractable and severely constrains the potential of test-time scaling.

To prevent the model from collapsing into an “answer earlier” strategy, we propose an over-turn masking technique whose objective is to avoid penalizing over-turn responses. The overall procedure is illustrated in Fig.~\ref{fig:overlength_mask}. Concretely, in addition to the rewards $r$ and advantages $A$ defined as in vanilla GRPO, we introduce a completion mask $M$ that indicates whether a response terminates successfully. We then compute masked advantages $A_i'=M_i \cdot A_i$, so that over-turn trajectories (with $M_i=0$) do not contribute negative learning signals. The modified objective, building on \eqref{eq:GRPO}, is summarized below, with the changes highlighted in red in the formula.
\begin{equation}
\begin{split}
    \mathcal{J}^{\textcolor{red}{over-turn}}_{GRPO}(\theta) & = \mathbb{E}_{[q \sim \mathcal{D}, \{o_i\}_{i=1}^G \sim \pi_{\theta_{old}}(\cdot|q)]}  \\
     \frac{1}{\textcolor{red}{\sum_i^{G}M_i}}\sum_{i=1}^G & \left(\min\left(\frac{\pi_\theta(o_i |q)}{\pi_{\theta_{old}}(o_i |q)} A_i \textcolor{red}{\cdot M_i}, \text{clip} \left( \frac{\pi_\theta(o_i |q)}{\pi_{\theta_{old}}(o_i |q)}, 1 - \epsilon, 1 + \epsilon \right)  A_i\textcolor{red}{\cdot M_i} \right) \right)
\end{split}
\label{eq:new_GRPO}
\end{equation}
\begin{equation}
\begin{split}
    \textcolor{red}{M_i = \mathds{1}\{|o_i| <= C_{context}\} \cdot \mathds{1}\{ \text{turn}(o_i) <= C_{turn}\}}.
\end{split}
\label{eq:new_advantage}
\end{equation}

Here, $|o_i|$ and $\text{turn}(o_i)$ denote the token length and the number of turns in response $o_i$, respectively. Moreover, because some responses are incomplete, we normalize the objective by the number of completed generations, $\sum_i^{G}M_i$, rather than by the total number of generations $G$.

With this technique, we mask out the loss for over-turn responses, thereby removing any implicit penalty. Notably, although we adopt a relatively small upper bound on the number of turns during training, test-time trajectories can extend to dozens of rounds, with accuracy improving monotonically. The proposed over-turn masking is thus essential for realizing the benefits of test-time scaling in the number of interaction turns, as illustrated in Fig.~\ref{fig:overlength_mask}.

\subsection{Inference}
\paragraph{Generation with Temperature} During inference, we observe that greedy decoding tends to produce \textit{repeated words or sentences}, likely because the effective context grows with the number of turns. To mitigate this issue, a simple yet effective method is to set the temperature to $1.0$, which introduces sufficient randomness to reduce repetition without substantially degrading coherence.

\section{Experiment}

\subsection{Experimental Setting}

\paragraph{Supervised Finetuning} During SFT, we use \texttt{Qwen2.5-VL-7B-Instruct} as the base model. Given the context-length constraints in multi-turn agentic interactions, we set the maximum pixel budget to 2M unless otherwise specified. We train on approximately $6,000$ cold-start samples for 3 epochs. The learning rate is set to $1\times10^{-5}$, and the global batch size is $32$.

\paragraph{Reinforcement Learning} For reinforcement learning, we follow DAPO~\citep{yu2025dapo} and adopt clip-higher, dynamic sampling, and a token-level policy loss to ensure stable training. We set the group size to $16$. By default, the upper and lower clip ratios are $0.30$ and $0.20$, respectively. The global batch size is $256$, with a mini-batch size of $32$. We use a constant learning rate of $1\times10^{-6}$. Neither KL regularization nor entropy regularization is applied. To maintain training efficiency, we cap the maximum number of turns at $6$ and set the maximum context length to $32$K tokens. We also implement asynchronous rollouts to accelerate training.

\paragraph{Dataset} For training, we use the VisualProbe training split. In addition, to preserve performance on simpler visual search cases, we randomly sample $8,000$ examples from DeepEyes-Datasets-47k~\citep{zheng2025deepeyes}. The test suites include VisualProbe-test, V\textsuperscript{*} Bench, HR-Bench, and MME-Realworld~\citep{zhang2024mme}.

\paragraph{Evaluation Metric} We find that single-run evaluation exhibits high variance and does not reliably reflect robustness due to sampling stochasticity. To mitigate this, we report the Avg@K metric: each problem is evaluated $K$ times with temperature set to $1.0$, and accuracy is computed by averaging across the $K$ responses.

\subsection{Main Result}

\begin{table}[t]
    \centering
    \caption{Performance comparisons among existing models and ours on visual search tasks. The sizes of all listed models are 7B. For VisualProbe and V\textsuperscript{*} Bench, we report Avg@32 to reduce variance caused by randomness. We report Avg@8 and Avg@1 for HR-Bench and MME-Realworld, respectively.}
    \label{table:main_result}
    \vspace{0.1cm}
    \tabcolsep=0.25cm
    {
        \begin{threeparttable}
        % \begin{footnotesize}
        \begin{tabular}{l|ccc|c|cc|c}
            \toprule
            \multirow{2}{*}{Model}
            & \multicolumn{3}{c|}{VisualProbe}
            & \multirow{2}{*}{V*}
            & \multicolumn{2}{c|}{HR-Bench} & \multirow{2}{*}{MME-Realworld} \\
            
            \specialrule{0em}{0pt}{1pt}
            \cline{2-4}
            \cline{6-7}
            \specialrule{0em}{0pt}{1pt}
            
            & hard & medium & easy &  & 4K & 8K \\
            \midrule
            % 示例数据行（可替换/删除）
            GPT-4o~\citep{gpt4o} & 11.2 & 15.4 & 47.5 & 65.2 & 62.0 & 58.3 & 45.2 \\
            % o3 & 24.2 & 20.2 & 52.6 \\

            \specialrule{0em}{0pt}{1pt}
            \hline
            \specialrule{0em}{0pt}{1pt}

            LLaVA-OneVision~\citep{li2024llava} & 13.4 & 12.5 & 36.2 & 70.9 & 61.2 & 54.0 & 57.4 \\
            Qwen2.5-VL-Instruct~\citep{bai2025qwen2.5} & 23.9 & 26.0 & 39.1 & 75.5 & 68.2 & 62.7 & 57.3 \\
            % NOTE: The performance of Qwen2.5-VL, LLaVA-OV, GPT4o on MME-Realworld is copied from its leaderboard.
            \specialrule{0em}{0pt}{1pt}
            \hline
            \specialrule{0em}{0pt}{1pt}
            SEAL$^{\dagger}$~\citep{wu2024v} & - & - & - & 75.4 & - & - & - \\
            DyFo$^{\dagger}$~\citep{li2025dyfo} & - & - & - & 81.2 & - & - & - \\
            % ZoomEye$^{\dagger}$~\citep{shen2024zoomeye} & - & - & - & 90.6 & 69.6 & 69.3 \\
            Chain-of-Focus$^{\dagger}$~\citep{zhang2025chain} & - & - & - & 88.0 & - & - & - \\
            Pixel Reasoner$^{\ddagger}$~\citep{su2025pixel} & 28.8 & 29.6 & 58.4 & 86.3 & 74.0 & 66.9 & 64.4 \\
            DeepEyes$^{\ddagger}$~\citep{zheng2025deepeyes} & 35.1 & 29.8 & 60.1 & 83.3 & 73.2 & 69.5 & 64.0 \\
            \rowcolor{Gray} Mini-o3 (Ours) & \textbf{48.0} & \textbf{50.4} & \textbf{67.0} & \textbf{88.2} & \textbf{77.5} & \textbf{73.3} & \textbf{65.5} \\
            
            % \rowcolor{Gray}Qwen2-72B-SFT + Step-DPO & 72B & \XSolidBrush & \Checkmark & 64.7 \textcolor[RGB]{20,160,20}{(+3.0)} & 93.9 \textcolor[RGB]{20,160,20}{(+1.0)} \\
            \bottomrule
        \end{tabular}
        % \end{footnotesize}
        \begin{tablenotes}
          \small
          \item $\;\;^{\dagger}$ The models only report the metric of Avg@1 and the model weights are not available.
          \item $\;\;^{\ddagger}$ Re-evaluated using its official model and evaluation code to yield the metric of Avg@32.
        \end{tablenotes}
        \end{threeparttable}
    }
    % \vspace{0.3cm}
% \vspace{-0.3cm}
\end{table}

The performance comparison between existing models and Mini-o3 on visual search tasks is presented in Table~\ref{table:main_result}. To ensure robust and convincing evaluation, we assess all models on VisualProbe, V\textsuperscript{*} Bench, and HR-Bench. Across all datasets, Mini-o3 achieves state-of-the-art performance, substantially outperforming other open-source baselines. We attribute these gains to Mini-o3’s ability to sustain more complicated and deeper reasoning trajectories.

% \xinlai{add a fig to show the average turns of mainstream models on different datasets}

\subsection{Ablation Study}

\begin{table}[t]
    %\footnotesize
    %\vspace{0.1cm}
    \centering
    \caption{Ablation study for main components of the method. Max pixels are set to $1$M. Upper limit on the number of turns is set to $6$ during training. Evaluations are made on VisualProbe test set.}
    \label{table:ablation}
    \tabcolsep=0.27cm
    {
        \begin{footnotesize}
        \begin{tabular}{c | c c c | c c c | c }
            \toprule
            
            ID & hard RL data & cold-start & over-turn & Hard & Medium & Easy & Avg. Turns (correct)\\
            
            \specialrule{0em}{0pt}{1pt}
            \hline
            \specialrule{0em}{0pt}{1pt}

            1 & & \Checkmark & \Checkmark & 35.8 & 46.4 & 66.7 & 4.8 \\
             
            2 & \Checkmark &  & \Checkmark & 25.4 & 18.7 & 57.3 & 1.0 \\
            
            3 & \Checkmark & \Checkmark & & 32.2 & 45.7 & 61.1 & 3.0 \\

            \specialrule{0em}{0pt}{1pt}
            \hline
            \specialrule{0em}{0pt}{1pt}
            
            4 & \Checkmark & \Checkmark & \Checkmark & \textbf{44.4} & \textbf{47.9} & \textbf{67.4} & \textbf{5.5}\\
            
            \bottomrule                                   
        \end{tabular}
        \end{footnotesize}
    }
    % \vspace{0.1cm}
% \vspace{-0.2cm}
\end{table}

\begin{table}[t]
    %\footnotesize
    %\vspace{0.1cm}
    \centering
    \caption{Ablation study on the values of max pixels. Evaluations are made on VisualProbe test set. Also, we calculate the average number of interaction turns among overall and correct trajectories.}
    \label{table:max_pixels}
    \tabcolsep=0.3cm
    {
        \begin{footnotesize}
        \begin{tabular}{c | c c c | c c }
            \toprule
            
            Max Pixels & Hard & Medium & Easy & Avg. Turns (All) &  Avg. Turns (Correct) \\
            
            \specialrule{0em}{0pt}{1pt}
            \hline
            \specialrule{0em}{0pt}{1pt}

            0.5M & 36.4 & 44.8 & 64.8 & \textbf{8.0} & \textbf{6.7} \\
             
            1M & 44.4 & 47.9 & \textbf{67.4} & 6.3 & 5.5 \\
            
            2M & \textbf{48.0} & \textbf{50.4} & 67.0 & 6.5 & 5.6 \\

            12M & 36.1 & 40.7 & 62.1 & 1.0 & 1.0\\
            % \specialrule{0em}{0pt}{1pt}
            % \hline
            % \specialrule{0em}{0pt}{1pt}
            
            \bottomrule                                   
        \end{tabular}
        \end{footnotesize}
    }
    % \vspace{0.1cm}
% \vspace{-0.2cm}
\end{table}

In this section, we present an extensive ablation study to quantify the contribution of each component in our method. The overall results are summarized in Table~\ref{table:ablation}. Unless otherwise specified, all experiments are conducted on the VisualProbe test set with the maximum pixel budget set to $1$M.

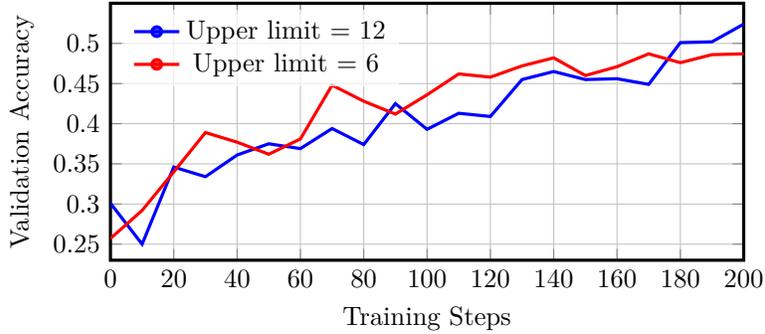
\begin{figure}[t]
\centering
\begin{tikzpicture}
  \begin{axis}[
    width=10cm, height=5cm,
    xlabel={Training Steps},
    ylabel={Validation Accuracy},
    xmin=0, xmax=200,
    ymin=0.23, ymax=0.55,
    xtick={0,20,40,60,80,100,120,140,160,180,200,220},
    ytick={0.25,0.30,0.35,0.40,0.45,0.50},
    grid=both,
    grid style={line width=.1pt, draw=gray!30},
    major grid style={line width=.2pt, draw=gray!50},
    legend style={at={(0.02,0.98)},anchor=north west,draw=none,fill=white,fill opacity=0.7,text opacity=1},
    cycle list name=color list,
    line width=1.2pt,
    mark=*,
    mark size=1.8pt
  ]
    % 数据 x 从 0 到 200，每 10 一点
    % Upper limit = 12
    \addplot+[blue] table[row sep=\\]{
      x   y \\
      0   0.301 \\
      10  0.250 \\
      20  0.346 \\
      30  0.334 \\
      40  0.361 \\
      50  0.375 \\
      60  0.369 \\
      70  0.394 \\
      80  0.374 \\
      90  0.425 \\
      100 0.393 \\
      110 0.413 \\
      120 0.409 \\
      130 0.455 \\
      140 0.465 \\
      150 0.455 \\
      160 0.456 \\
      170 0.449 \\
      180 0.501 \\
      190 0.502 \\
      200 0.524 \\
      % 210 0.513 \\
      % 220 0.539 \\
    };
    \addlegendentry{Upper limit = 12}
    % Upper limit = 6
    \addplot+[red] table[row sep=\\]{
      x   y \\
      0   0.257 \\
      10  0.292 \\
      20  0.340 \\
      30  0.389 \\
      40  0.377 \\
      50  0.362 \\
      60  0.381 \\
      70  0.448 \\
      80  0.428 \\
      90  0.412 \\
      100 0.436 \\
      110 0.462 \\
      120 0.458 \\
      130 0.472 \\
      140 0.482 \\
      150 0.460 \\
      160 0.471 \\
      170 0.487 \\
      180 0.476 \\
      190 0.486 \\
      200 0.487 \\
      % 210 0.467 \\
      % 220 0.468 \\
    };
    \addlegendentry{Upper limit = 6}
  \end{axis}
\end{tikzpicture}
\caption{Accuracy on VisualProbe-Hard during the training progress. The upper limit of the number of turns is set to $6$ and $12$, respectively.}
\label{fig:ablation_upper_limit}
\end{figure}

\paragraph{Hard RL Data} We compare experiments 1 and 4 in Table~\ref{table:ablation}. Removing the hard RL data leads to a performance decrease of approximately 8.6 points on VisualProbe-Hard, indicating that challenging RL samples are crucial for encouraging complex reasoning trajectories.

\paragraph{Cold-start SFT} To assess the necessity of cold-start SFT, we contrast experiments 2 and 4 in Table~\ref{table:ablation}. The results show that cold-start SFT is essential for multi-turn tool use: performance collapses without it. We hypothesize that the base model lacks exposure to multi-turn agentic trajectories during pre-training or instruction tuning, and cold-start SFT serves as a pivotal initialization.

\paragraph{Over-turn Masking} A comparison between experiments 3 and 4 in Table~\ref{table:ablation} demonstrates that over-turn masking benefits reinforcement learning, particularly in multi-turn settings. It offers two main advantages. First, it stabilizes training by avoiding incorrect penalization of truncated responses whose correctness is inherently uncertain. Second, it enables test-time turn scaling and unlocks strong performance on highly challenging tasks that require substantially more turns than the training-time upper bound. This trend is further corroborated in Fig.~\ref{fig:turn_dist}.

\paragraph{Max Pixels} Table~\ref{table:max_pixels} evaluates different maximum pixel budgets. We observe that both overly large and overly small settings are suboptimal. An excessively large budget induces premature “early stopping”, reducing the number of interaction turns and limiting iterative refinement. Conversely, a small budget increases perceptual hallucinations. We also report the average number of interaction turns in the same table, which highlights a trade-off between perceptual accuracy and interaction depth. Optimal overall performance is achieved by appropriately tuning the max-pixel budget.

\paragraph{Upper Limit on Turns During Training} To quantify the effect of a larger interaction-turn budget during training, we track the accuracy on VisualProbe-Hard over the course of training and compare budgets of $6$ and $12$ turns in Fig.~\ref{fig:ablation_upper_limit}. A lower budget leads to faster initial convergence, but the performance plateaus after approximately 150 steps. In contrast, a higher turn budget attains a superior performance ceiling, albeit with slower convergence.

\section{Conclusion}

In this work, we investigate multi-turn image-based tool use for Vision-Language Models (VLMs). To address challenging visual search problems that demand iterative exploration and trial-and-error, we introduce Mini-o3, a model capable of producing diverse reasoning patterns and deep chains of thought. Its trajectories scale to tens of turns, during which accuracy continues to improve, yielding substantial gains over prior models on multiple visual search benchmarks.
To enable these capabilities, we develop a three-pronged approach. First, we construct VisualProbe, a challenging visual search dataset comprising both training and evaluation tasks. Second, we devise a simple yet effective pipeline for collecting cold-start data by leveraging the in-context learning ability of an existing VLM. Third, we enhance vanilla GRPO with an over-turn masking strategy that prevents undue penalties on responses that exceed the training budget on turns. This modification facilitates test-time turn scaling and enables the solution of particularly difficult problems. We believe this recipe offers practical guidance for reinforcement learning and the development of multimodal models with multi-turn interactions.

% We sincerely acknowledge that 

\clearpage

\bibliographystyle{plainnat}
\bibliography{main}

\clearpage

\beginappendix

\section{More illustrations of multi-turn trajectories}

\begin{figure}[H]
\centering
\includegraphics[width=0.99\linewidth]{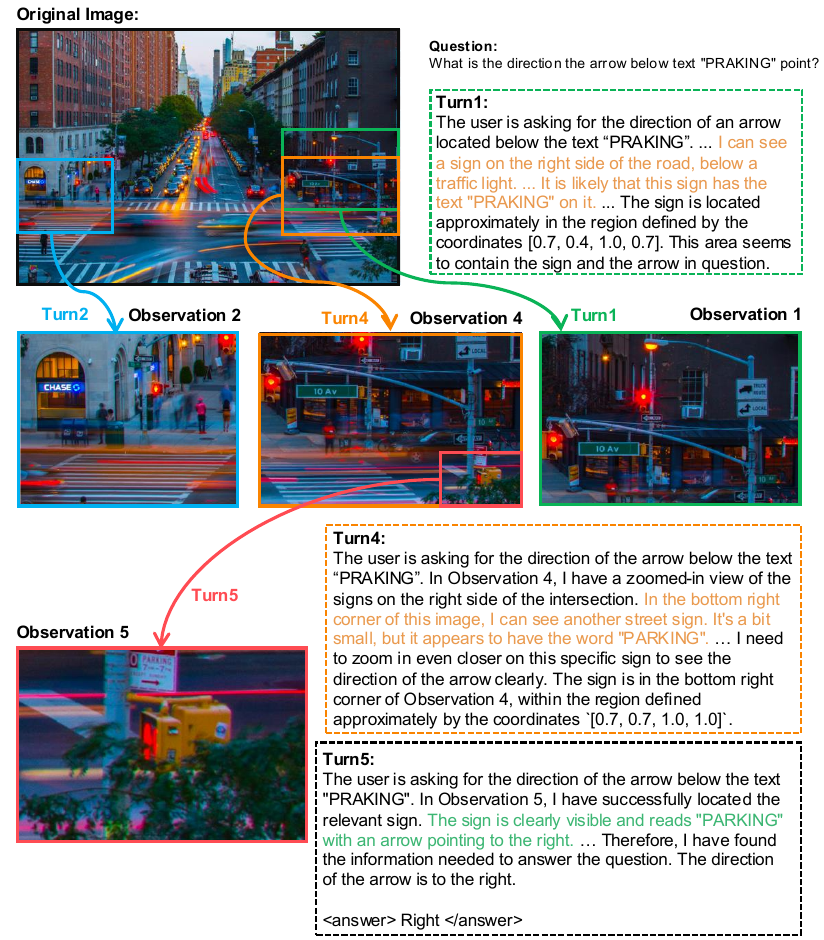}
\vspace{-0.1cm}
\caption{An example of visual search in a urban intersection generated by Mini‑o3. The scene shows a busy city street with multiple signs and traffic elements. Our model conducts multi-turn reasoning—progressive zoom-in, hypothesis revision, and backtracking—to identify the direction the arrow is pointing.}
\label{fig:appendix_demo_0}
\end{figure}

\begin{figure}[H]
\centering
\includegraphics[width=0.99\linewidth]{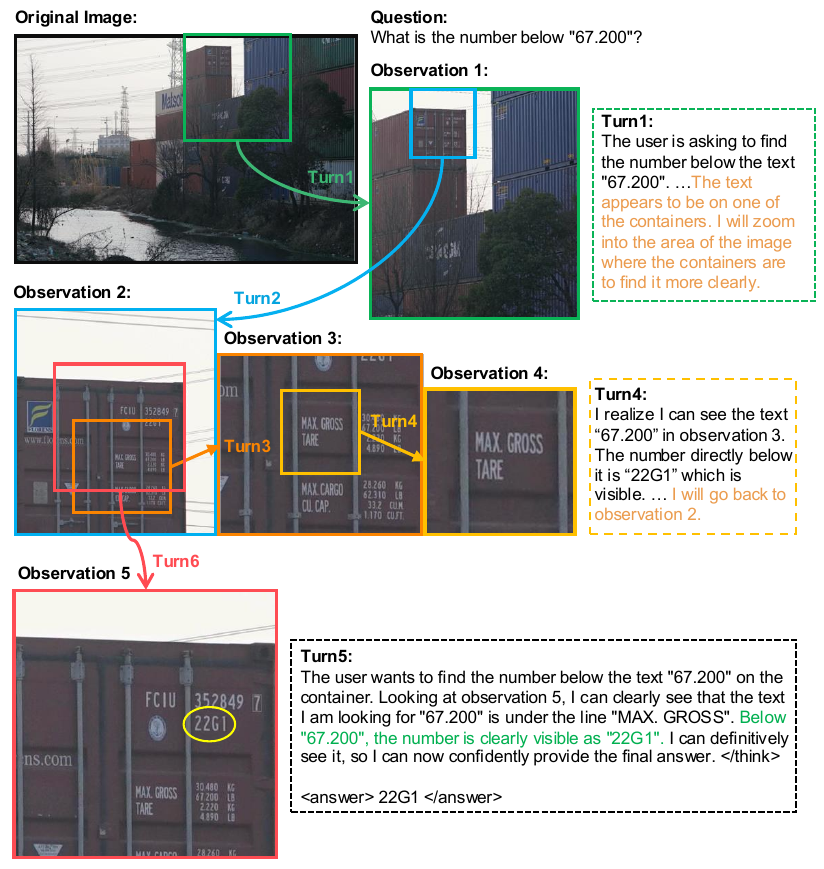}
\vspace{-0.1cm}
\caption{An example of visual search in a container yard generated by Mini‑o3. The scene depicts stacked shipping containers with dense text markings. Our model performs multi‑turn reasoning—targeted zoom-ins, cross-checking across observations, and corrective backtracking—to locate the string ``67.200" and read the number directly beneath it. It outputs the correct value, ``22G1", demonstrating precise text localization and robust step-by-step verification in a cluttered setting.}
\label{fig:appendix_demo_1}
\end{figure}

\begin{figure}[H]
\centering
\includegraphics[width=0.99\linewidth]{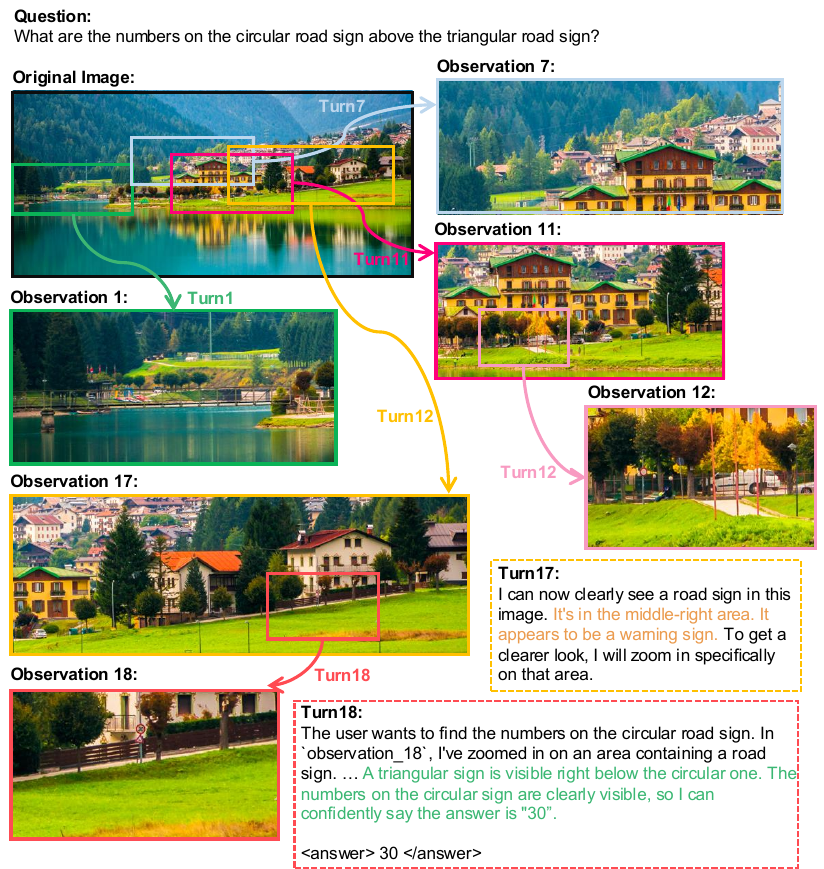}
\vspace{-0.1cm}
\caption{An example of visual search in a lakeside village generated by Mini‑o3. Our model performs multi‑turn reasoning—coarse‑to‑fine zooming, refocusing, and verification across observations—to localize a circular road sign above a triangular warning sign. Mini-o3 ultimately recognizes the digits ``30" on the sign after 18 reasoning turns.}
\label{fig:appendix_demo_2}
\end{figure}

\end{document}